\documentclass{article}
\usepackage{iclr2019_conference,times}

\usepackage[utf8]{inputenc} 
\usepackage[T1]{fontenc}    
\usepackage{hyperref}       
\usepackage{breakurl} 
\usepackage{url}            

\usepackage{booktabs}       
\usepackage{amsfonts}       
\usepackage{nicefrac}       
\usepackage{microtype}      
\usepackage{placeins} 
\usepackage{verbatim} 
\usepackage{glossaries} 

\usepackage{graphicx,color}
\usepackage{colortbl} 

\usepackage{color,soul}

\usepackage[font=footnotesize]{subcaption} 
\usepackage[inline]{enumitem}

\usepackage{multirow}
\usepackage{array}

\PassOptionsToPackage{numbers, compress}{natbib}

\title{Automatic generation of object shapes with desired functionalities} 

\author{
  Mihai Andries \\
  Institute for Systems and Robotics\\
  Instituto Superior Técnico \\
  Lisbon, Portugal \\
  \texttt{mandries@isr.tecnico.ulisboa.pt}
  \And
  Atabak Dehban \\
  Institute for Systems and Robotics\\
  Instituto Superior Técnico \& \\
  Champalimaud Centre
  for the Unknown \\
  Lisbon, Portugal \\
  \texttt{adehban@isr.tecnico.ulisboa.pt}
  \And
  José Santos-Victor \\
  Institute for Systems and Robotics\\
  Instituto Superior Técnico\\
  Lisbon, Portugal \\
  \texttt{jasv@isr.tecnico.ulisboa.pt}
}

\newglossaryentry{VAE}
{
  name={VAE},
  description={Variational AutoEncoder},
  first={\glsentrydesc{VAE} (\glsentrytext{VAE})},
  plural={VAEs},
  descriptionplural={variational autoencoders},
  firstplural={\glsentrydescplural{VAE} (\glsentryplural{VAE})}
}

\newglossaryentry{PDF}
{
  name={PDF},
  description={Probability Density Function},
  first={\glsentrydesc{PDF} (\glsentrytext{PDF})},
  plural={PDFs},
  descriptionplural={Probability Density Functions},
  firstplural={\glsentrydescplural{PDF} (\glsentryplural{PDFs})}
}

\newglossaryentry{DARPA}
{
  name={DARPA},
  description={the Defense Advanced Research Projects Agency},
  first={\glsentrydesc{DARPA} (\glsentrytext{DARPA})}
}

\newglossaryentry{CAD}
{
  name={CAD},
  description={Computer-Aided Design},
  first={\glsentrydesc{CAD} (\glsentrytext{CAD})}
}

\newacronym{SDF}{SDF}{Spatial Data File}
\newacronym{FCT}{FCT}{Fundação para a Ciência e a Tecnologia}
\newacronym{SURF}{SURF}{Speeded up robust features}
\newacronym{SIFT}{SIFT}{Scale-invariant feature transform}
\newacronym{KL}{KL}{Kullback–Leibler}
\newacronym{GAN}{GAN}{Generative Adversarial Network}
\newacronym{ReLU}{ReLU}{rectified linear unit}
\newacronym{IST-ID}{IST-ID}{Association of Instituto Superior Técnico for Research and Development} 
\makeglossaries

\begin{document}

\maketitle

\begin{abstract}

    3D objects (artefacts) are made to fulfill functions.
    Designing an object often starts with defining a list of functionalities that it should provide, also known as \emph{functional requirements}.
    Today, the design of 3D object models
        is still a slow and largely artisanal activity,
        with few \gls{CAD} tools
        existing to aid the exploration of the design solution space.
    %
    To accelerate the design process, 
        we introduce an algorithm 
        for generating object shapes with desired functionalities.
    Following the concept of \emph{form follows function},
        we assume that existing object shapes were rationally chosen
        to provide desired functionalities.
    %
    First, we use an artificial neural network to
        learn a function-to-form mapping
        by analysing a dataset of 
        objects labeled with their functionalities.
    Then, we combine forms providing one or more desired functions,
        generating an object shape that is expected to provide all of them. 
    Finally, we verify in simulation 
        whether the generated object possesses the desired functionalities,
        by defining and executing functionality tests on it.
\end{abstract}

\section{Motivation}

    Design cycles of industrial products are lengthy, as they usually involve thousands of decisions on the form of the product that will implement the desired functionalities.
    Despite efforts in the last two decades to accelerate the workflow using \gls{CAD} techniques \cite{kurtoglu2007_phdthesiscomputational,Autodesk_Dreamcatcher}, most of the design process is still done manually. This is because all the proposed automation techniques (adaptive design, generative design, etc.) require
    input in the form of human decisions or expert knowledge, which is generally difficult to obtain and formulate.
    %
    In an attempt to solve this pertinent problem,
    \gls{DARPA} launched in 2017 the \emph{Fundamental Design} call for research projects on conceptual design of mechanical systems,
    that would enable the generation of novel design configurations \cite{DARPA_Fun_Design}.

    This paper also has another motivation stemming from robotics.
    Traditionally, research in autonomous robots deals with the problem
        of recognising affordances of objects in the environment:
        i.e. given an object, what actions does is afford to do?
        Given a shape, what are its functionalities?
    This paper addresses the inverse problem: given a list of functionalities, what shape would provide all of them?

\begin{figure}[t]
        \centering
            \begin{subfigure}[t]{.32\linewidth}
                \centering
                \includegraphics[height=0.55\linewidth]{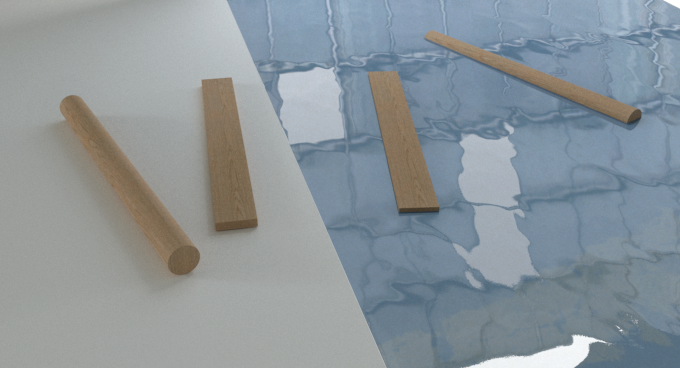}
              \caption{\emph{Wooden} beams have the \emph{float-ability} affordance.}
              \label{subfig:wooden_beam}
            \end{subfigure}
            ~
            \begin{subfigure}[t]{.32\linewidth}
                \centering
                \includegraphics[height=0.55\linewidth]{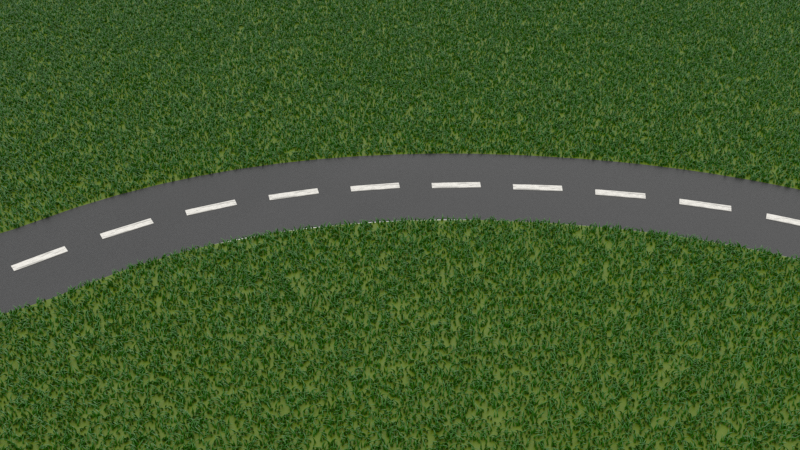}
              \caption{\emph{Flat} roads have the \emph{traverse-ability} affordance.}
              \label{subfig:road}
            \end{subfigure}
            ~
            \begin{subfigure}[t]{.32\linewidth}
                \centering
                \includegraphics[height=0.55\linewidth]{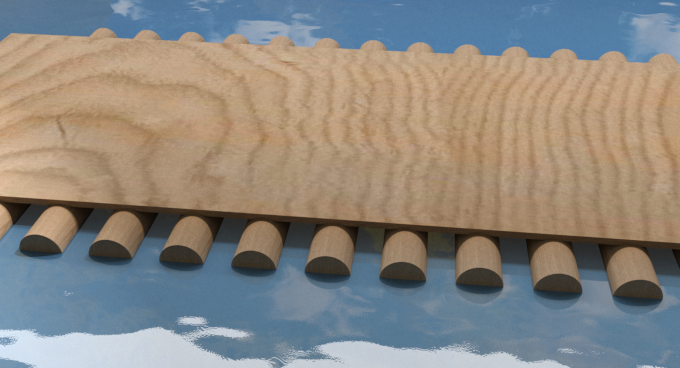}
              \caption
              {
                \emph{Flat wooden} roads (pontoon bridges) offer both 
                \emph{float-ability} and \emph{traverse-ability}.
              }
              \label{subfig:pontoon_bridge}
            \end{subfigure}
        \caption
        {
            The features that describe
            (\ref{subfig:wooden_beam}) wooden beams and
            (\ref{subfig:road}) flat roads
            can be combined, to obtain an object design that possesses both \emph{float-ability} and \emph{traverse-ability}:
            (\ref{subfig:pontoon_bridge}) a pontoon bridge.
        }
        \label{fig:pontoon_bridge_example}
    \end{figure}

    This paper presents a method and an architecture
        for automatic generation of object shapes with desired functionalities.
    It does so by autonomously learning mappings from object form to function, and then applies this knowledge to conceive new object forms that satisfy  given functional requirements.
    In a sense, this method performs
        \emph{functionality arithmetic}
        through manipulation of object form.
    Fig.~\ref{fig:pontoon_bridge_example} illustrates the concept: 
    combine \emph{features} describing two different objects
    to create another object possessing the \emph{functionalities} of both initial objects.
    A quick skim through the other figures of this paper may help the reader understand what we are talking about.
    A second contribution is the use of experiments to verify the presence of affordances in the generated object shapes using a physics simulator, both by defining explicit tests in the simulator, and by using state-of-art affordance detectors.
    To summarise, this paper has three contributions: \begin{enumerate*}[label=(\roman*)]
        \item a novel method for extracting and combining functional essences of object shapes,
        \item design and execution of specific validation tests for the presence of desired affordances,
        \item a novel network architecture specialised for modeling 3D shapes.
    \end{enumerate*}

    %

    This remainder of the paper is organised as follows.
    Section~\ref{sec:related_work} presents an overview of the related work in 
    object design, shape descriptors, and object affordances.
    Section~\ref{sec:methodology} describes our methodology,
        detailing the envisioned workflow for using this technology,
        and details regarding the architecture of the network.
    It also describes the operators employed for object form manipulation.
    In Section~\ref{sec:evaluation_discussion}
        we discuss the obtained results and
        describe the drawbacks of the method at its current state.
    Finally, in Section~\ref{sec:conclusion_future_work}
    we draw a conclusion and detail the opportunities for future work.
    In this paper we will use the terms \emph{affordance} and \emph{functionality} interchangeably.

\section{Related literature}
\label{sec:related_work}

    The literature review is organised in three sections, detailing the state-of-the-art in the three fields
    at the intersection of which this study finds itself:
    object design,
    object shape descriptors (for manipulation of object forms),
    and
    learning of object affordances (for relating object forms to functionalities).

    \subsection{Object design}

    The idea of
    getting inspiration from previous designs
    when conceiving a new object
    is not new, and appears under names such as Analogical reasoning, and Design reuse.
    A standard practice in design is to consult \emph{knowledge ontologies}
    \protect{\citep{bryant2005concept_functional_design,kurtoglu_2009_eletromech_design_function_form,bhatt2012ontological}}
    that contain function-to-form mappings \citep{umeda_1997_functional_design_585103,kurtoglu2007_phdthesiscomputational}.
    However, the knowledge acquisition required to populate such ontologies 
    involves a (non-automated) process known as \emph{functional decomposition},
    in which a human analyses existing objects by disassembling them into components and noting the functionality provided by each component.
    A related review on object functionality inference from shape information is presented in \citep{hu2018_functionality_representations_shape_analysis}.

    Recently, generative design emerged as an automated technique for exploring the space of 3D object shapes \citep{Autodesk_Dreamcatcher} using genetic algorithms.
    It formulates the shape search as an optimisation problem, requiring an initial solution, a definition of parameters to optimise, and rules for exploring the search space.
    However, it is far from trivial to identify rules for the intelligent exploration of the shape space, that would provide results in reasonable time.
    In a similar context of generative design, \cite{Umetani_2017_autoenc_cars} employed an AutoEncoder to explore the space of car shapes.

    \subsection{Object shape descriptors}

    Object shape descriptions serve two purposes:
        (1) they contain extracted object shape features, which are used to study the form-to-function relationship, and
        (2) they serve as basis for the reconstruction of 3D object models.
    State-of-the-art techniques for automatically extracting object features are practically all based on Neural Networks, typically Convolutional Neural Networks or Auto-Encoders \citep{girdhar_2016_Generative_Vector_repr_objects}, which have replaced the methods based on hand-crafted features like \gls{SIFT} or \gls{SURF}.

    In order to 
    generate 3D shapes from descriptions, modern techniques also employ Neural Network approaches: Auto-Encoders \citep{girdhar_2016_Generative_Vector_repr_objects} and Generative Adversarial Networks \citep{wu_GAN_NIPS2016_6096}, which learn a mapping from a low-dimensional probabilistic latent space to the space of 3D objects, allowing to explore the 3D object manifold.
    In this study, we used a \gls{VAE} \citep{vae_kingma_2013_arXiv1312.6114K,rezende_vae_2014_2014arXiv1401.4082J}
        to 
        both extract features describing 3D objects, and
        reconstruct their 3D shapes when given such a description.

    \subsection{Object functionalities as object affordances}

    A field of research
        that also focuses on linking objects with their functionalities
        is that of \emph{affordance learning}.
    It is based on the notion of \emph{affordance}
        that defines an action that an object provides (or affords) to an agent
        \citep{gibsonj_affordances_1977}.
    In the context of this paper,
        we are interested in approaches that map
        object features to corresponding object affordances (or functionalities).
    A common approach is to
        extract image regions (from RGB-D frames) with specific properties
        and tag them with corresponding affordance labels.
    An overview of machine learning approaches for detecting affordances of
        tools in 3D visual data is available in 
        \citep{ciocodeica2016machine}.
    Recent reviews on 
        affordances in machine learning for cognitive robotics
        include
        \citep{Jamone2016,min_2016_affordance_review,zech_affordance_review_2017}.

    This paper
        introduces a method to automatically learn shape descriptors
        and extract a form-to-function mapping,
        which is then employed
        to generate new objects with desired functionalities.
    The novelty lies in the use
        of what we call \emph{functionality arithmetic} 
        (operations on object functionalities)
        through manipulation of corresponding forms 
        in a feature space. 
    This is an application of the concept \emph{form follows function}
        in an automated design setting.

\section{Methodology}
\label{sec:methodology}

    The starting point for this research
        was the hypothesis that
        object functionalities
        arise due to features that those objects possess.
    Therefore, if we intend to create an
        object with a desired set of functionalities,
        then it should possess corresponding features
        providing these functionalities.

    In this section we describe
        our workflow for
        object generation
        (Section~\ref{sec:method_workflow}),
        technical details on the employed neural network architecture
        and its training (Section~\ref{sec:neural_network_architecture}),
        and the \emph{functionality arithmetics} operators that we used for
        generating shapes with desired functionalities
        (Section~\ref{sec:operators}).

    \subsection{Proposed workflow}
    \label{sec:method_workflow}

    For employing the proposed object generation method,
    we suggest a workflow composed of two phases:
    (1) learning phase, in which a neural network
        is trained to generate 
        feature-based
        representations of objects
        and to faithfully reconstruct objects using this representation,
    and
    (2) request phase, in which a user requests the generation of a novel object
    with some desired functionalities among those 
    present 
    in the traning dataset of
    affordance-labeled 
    objects.
    The algorithm would then pick object categories providing those functionalities,
    extract the shape features responsible for providing those functionalities (generating the form-to-function mapping), 
    and combine them to generate a feature description of a new object.
    This description would then be used to generate a 3D model of the desired object.

    \subsection{Neural network Architecture}
    \label{sec:neural_network_architecture}

    To come up with an automatic method for describing
        the features of objects,
        we employ a \gls{VAE}
        that we train on the ModelNet 40 dataset of common household objects
         \citep{3d_shapenets_modelnet_2015_7298801}.
    This dataset contains 3D models of
        bathtubs, beds, chairs, desks,
        dressers, monitors, night stands, sofas,
        tables, toilets, etc.
        (see examples in Fig.~\ref{fig:origs_and_recons}).
    For processing,
        we convert the samples in the dataset
        from OFF to binary voxelgrid (BINVOX) format \citep{binvox,nooruddin03},
        obtaining exact voxelgrid models
        centered in a volume of dimension 64x64x64 voxels.
    We augment the dataset by rotating the voxelgrid models
        by 90, 180 and 270 degrees around their vertical axis.

\begin{figure}[t]
        \centering
        \begin{subfigure}[t]{0.95\linewidth}
            \centering
            \includegraphics[width=\linewidth]{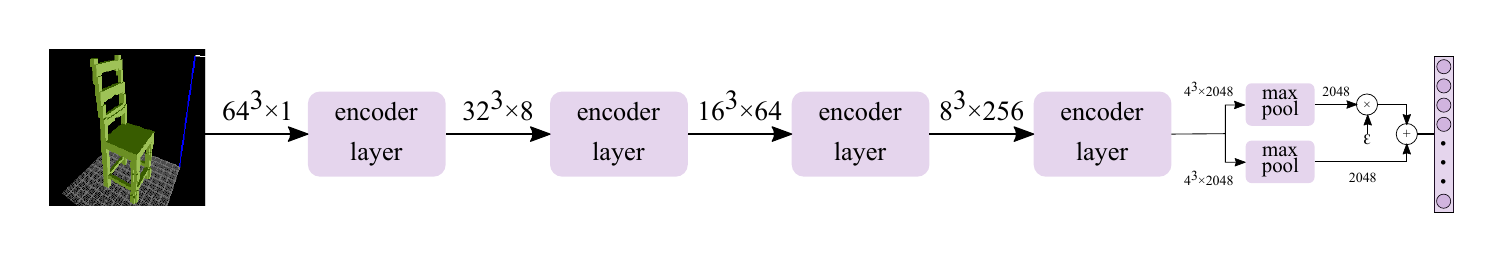}
            \vspace{-2em}
          \caption{Encoder architecture}
          \label{subfig:encoder}
        \end{subfigure}
        \\ 
        \begin{subfigure}[t]{0.95\linewidth}
            \centering
            \includegraphics[width=\linewidth]{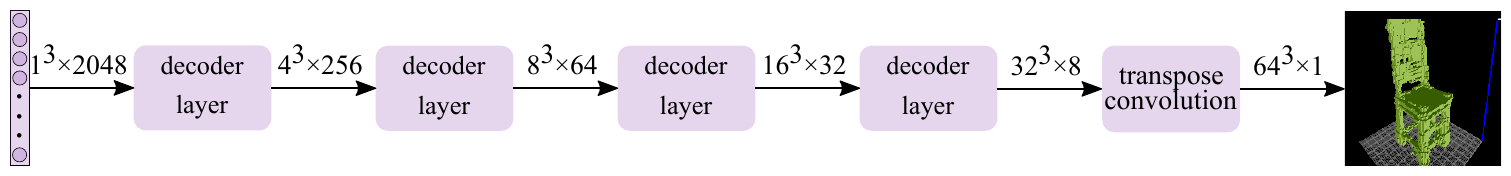}
            \caption{Decoder architecture}
            \label{subfig:decoder}
        \end{subfigure}
        \caption
        {The architectures of the~(\subref{subfig:encoder}) encoder and ~(\subref{subfig:decoder}) decoder.
        }
        \label{fig:architectures}
    \end{figure}

\begin{figure}[t]
    \centering
        \begin{subfigure}[t]{0.45\linewidth}
            \centering
            \includegraphics[width=\textwidth]{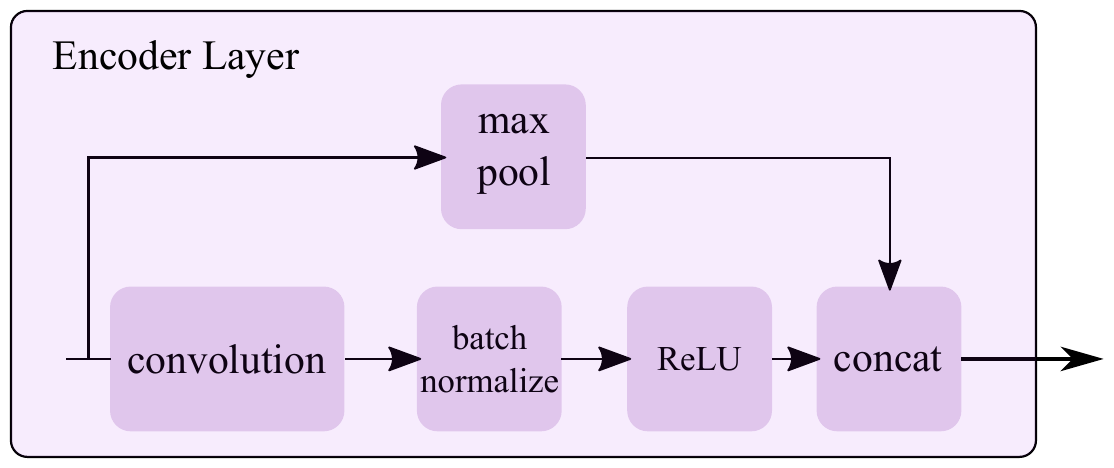}
          \caption{Encoder layer}
          \label{subfig:enc-layer}
        \end{subfigure}
        ~
        \begin{subfigure}[t]{0.45\linewidth}
            \centering
            \includegraphics[width=\textwidth]{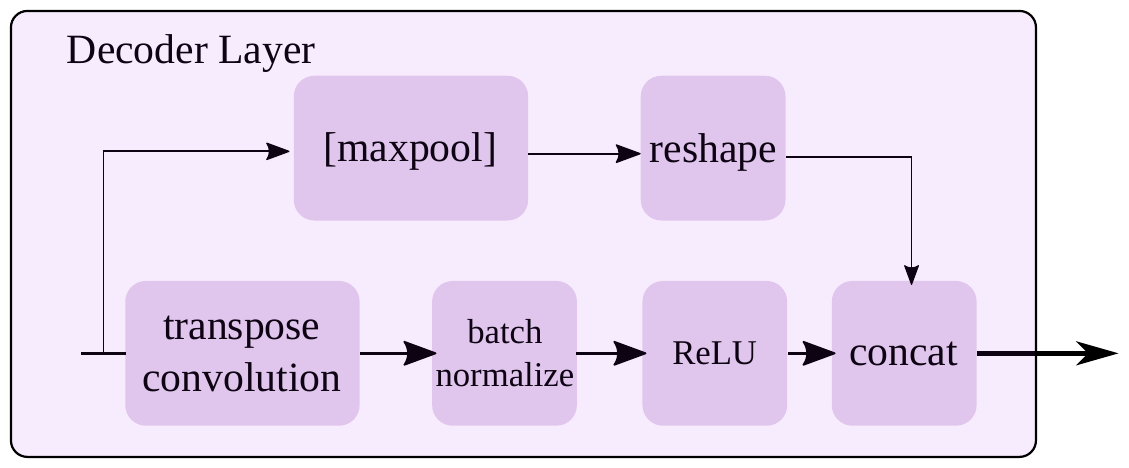}
            \caption{Decoder layer}
            \label{subfig:dec-layer}
        \end{subfigure}
        \caption{Schematics of the building blocks of the (\subref{subfig:enc-layer}) encoder layer and (\subref{subfig:dec-layer}) decoder layer.
        }
        \label{fig:building-blocks}
    \end{figure}

    The network architecture is shown in Figure \ref{fig:architectures}.
    The inputs are cubes of size $64 \times 64 \times 64$,
        which is identical to the dimensions of the reconstructed outputs.
    Both the encoder and the decoder employ convolutional layers,
        interspersed  with \gls{ReLU} non-linearities and batch normalization~\citep{ioffe2015batch} operations.
    Inspired by DenseNet architecture~\citep{huang2017densely}, we have stacked the outputs of activation layers throughout the encoder and decoder layers. However, unlike the bottleneck layers of~\citep{huang2017densely}, we have used max-pool operations~(and reshape operation, in the case of the decoder) to align the shapes~(see Figure~\ref{fig:building-blocks}).
    This was motivated by the fact that a significant bottleneck already exists in the latent variable layer.

    The last layer of the encoder performs a reduce-max operation,
        in order to generate the means and variances for the
        gaussian distributions that model each of the variables of the latent vector.
    The \gls{VAE} employs a latent vector of size $2^{11}$ (i.e. 2048)
        latent variables,
        which serves both as a bottleneck
        and as container of the object description.
    %
    We use a \gls{VAE} loss to train the network, which is composed of two parts: weighted binary cross-entropy (the reconstruction loss) and \gls{KL} divergence with a non-informative prior (a Gaussian with zero mean and unit variance) which is a regularisation loss.
    For a single example, the (non-weighted) reconstruction loss is computed as follows:
    \begin{equation}
        \sum - \big( x \cdot log(x') + (1 - x) \cdot log (1 - x') \big)
    \end{equation}
    where $x$ is the data,
    and $x'$ is the reconstruction.
    %
    To improve training speed,
        we employ a weighted cost function,
        that penalises proportionately more the network
        for errors in reconstructing full voxels
        than for errors in reconstructing empty voxels:
        \begin{equation}
            %
            \mathbb{E}_{q(z|x)}[log \ p(x|z)] =
            \sum - \big(\alpha \cdot x \cdot log(x') +
            (1 - x) \cdot log(1 - x') \big)
            %
        \end{equation}
        where
        $\alpha$ is the weight factor for the filled voxels,
        $log(.)$ is applied element-wise,
        and the summation is over the whole volume.
    %
    This is useful, since on average most of the reconstructed volume is empty,
        while the objects occupy only $\approx 4\%$ of all the voxels. 
    This allows to avoid the local minimum trap
        at the beginning of training,
        when the network prefers to reconstruct only empty volumes.
    We empirically set this weight
        to a value of $\alpha = 10$.
    The regularisation loss is computed as:
    \begin{equation}
        \label{eq:reg_loss}
        D_{KL} (q(z|x) || p(z)) = \frac{1}{2} \sum_{j=1}^{J} (1+log (\sigma_j^2) - \mu_j^2 - \sigma_j^2)
    \end{equation}
    where $J$ is the number of latent variables ($2048 = 2^{11}$ in our case),
    and $\mu_j$ and $\sigma_j$ are the parameters of the posterior of the latent variables (i.e. probability of the latent vector given the observation of a single data point: $q(z | x)$).
    %
    To improve the quality of
        reconstructions,
        we use a modified version of Eq.~\ref{eq:reg_loss}, referred to as \emph{Soft free bits} \citep{vae_lossy_2017_softbits_DBLP:journals/corr/ChenKSDDSSA16}
        that encourages the network to use all the capacity of
        its latent layer,
        by ensuring that
        each latent variable encodes at least $\lambda$ bits of information
        (with $\lambda$ given by the user):
    \begin{equation}
        \alpha_{SoftFreeBits}(x; \theta) =
        \mathbb{E}_{q(z|x)}[log \ p(x|z)] - \gamma D_{KL} (q(z|x) || p(z))
    \end{equation}
    with $0 < \gamma \leq 1 $,
    which is a parameter that increases the influence
    of the regularisation term for a specific latent variable
    if that variable does not contain $\lambda$ bits of information. More specifically, if the \gls{KL} divergence of any component of the posterior distribution with the prior goes below $\lambda$, $\gamma$ for that component is reduced such that by minimizing its influence on the total loss, the network can increase the amount of encoded information in that component without increasing the loss.
    On the other hand, if at least $\lambda$ bits of information are contained in a component, the corresponding component of $\gamma$ increases to one, so that the objective function better approximates the \emph{evidence lower-bound}. In this work, $\gamma$ starts at the value~$0.01$ and $\lambda$ is kept at $0.1$ during training. All other parameters are adopted from~\citep{vae_lossy_2017_softbits_DBLP:journals/corr/ChenKSDDSSA16}.
    \FloatBarrier

    \subsection{Operators}
    \label{sec:operators}

    In this section we describe the operators that we employed
        for manipulating object forms.
    Section~\ref{sec:extract_functional_essence} will describe the extraction of 
    \emph{functional essence} of a class of objects, which is the set of features that provides the functionalities of that class.
    Section~\ref{sec:object_combination} will describe how we combine
    two object descriptions into a single new one, that is expected to have the functionalities of both input objects.

    \subsubsection{Extract the \emph{functional essence} of a class of objects}
    \label{sec:extract_functional_essence}

        Every class of objects
            possesses a set of functionalities that defines it.
        From a \emph{form follows function} perspective,
            all objects samples contained in a class
            share a set of features that provide its set of functionalities.
        We call this set of features the \emph{functional essence}
            of a class of objects.
        To extract this \emph{functional essence},
            we compute the feature
            representation for each object in the class
            (i.e. the means and variances of variables composing the object description),
            and then average them, obtaining a single latent vector
            describing the sought \emph{functional essence}.
        This functional essence of an object class can then be visualised
        by inputting the obtained feature (latent-vector) description
        into the decoder trained to reconstruct 3D volumes.
        Fig.~\ref{fig:functional_essences} illustrates some results obtained using this method.

        Later,
        we assign an importance value to each
        latent variable composing the \emph{functional essence} of a class.
        We do this by computing the \gls{KL} divergence between the
        \gls{PDF} of these variables with the \gls{PDF} of
        (1) variables describing a void volume, and
        (2) a non-informative distribution
        of independent Gaussians with 0 mean and unit variance (called \emph{prior}).
        Both of these \gls{KL} divergences are normalised, so as to have unit norm.
        Then, an \emph{importance vector} is defined as the weighted sum of
        the normalized \gls{KL} divergences with a void
        and a non-informative prior distribution,
        with the corresponding weights
        $w_{void} = 2/3$ and $w_{prior} = 1/3$
        chosen empirically.

    \subsubsection{Combine the \emph{functional essences} of two different classes of objects}
    \label{sec:object_combination}

    In order to combine two object descriptions (i.e. two latent vectors containing these descriptions), we need to identify which of the variables in each vector are important for encoding the object shape. 
    In a degenerate case, if all the variables are critical for encoding the object shape, then their values cannot be changed, and therefore the object cannot be combined with another one (or a conflict resolution function must be devised).
    The hypothesis is that not all the variables are critical for representing the object shape, meaning that some variables' values can be neglected
    when combining two object descriptions.
    We identify which variables are important for an object description
        using the \emph{importance vector} method described above in Section~\ref{sec:extract_functional_essence}.

    The combination of two object descriptions
        is guided by their corresponding \emph{importance vectors}.
    For simplicity,
        we describe the combination as being made between two object descriptions,
        although the method is applicable to any number of objects.
    One object serves as a \emph{base object},
        from which are taken the initial values of the latent variables' distributions for the \emph{combined object} description.
    The other object serves as \emph{top object},
        whose latent variables' distributions
        are combined with those of the \emph{base object}
        according to the rules described in Table~\ref{table:combination_rules}.

    Four cases appear when combining two latent vector descriptions of objects,
        as seen in Table~\ref{table:combination_rules}.
    These rules can be resumed as follows:
        if both variable distributions are important then average them (case 4 in the table),
        if only one is important then keep the important one (cases 2 and 3 in the table),
        else keep the base values (case 1 in the table).

		\begin{table}[t]
			\centering
                \caption
                {
                   Interaction cases between latent variables contained in the descriptions of two different objects ($Obj_{base}, Obj_{top}$), which appear when attempting to combine them.
                }
                \label{table:combination_rules}
                \vspace{0.5em}
                \small
				\begin{tabular}[t]{m{0.05cm} m{3.90cm}  m{3.9cm}  m{4.5cm}}
	      \toprule
				\textbf{\#} &
                \textbf{Latent variable from $Obj_{base}$} &
                \textbf{Latent variable from $Obj_{top}$} &
                \textbf{Latent variable from $Obj_{combined}$} \\
	      \midrule
				1 & non-important & non-important &  value of base object \\
				2 & non-important & important & value of important variable \\
                3 & important & non-important & value of important variable \\
	            4 & important & important &
                average of the two values
				\end{tabular}
	      \normalsize
	  \end{table}


        \FloatBarrier

\section{Results and Discussion}
\label{sec:evaluation_discussion}

    In this section we provide
    our results on the
    (a) capacity of the \gls{VAE} to describe and reconstruct objects,
    (b) extraction of functional essences for different categories of objects,
    (c) generation of novel objects through the combination of feature representations of object classes containing desired functionalities, and
    (d) affordance testing for the generated objects.
    At the end of this section,
    we discuss the limitations of the proposed method.

    \subsection{Object representation and reconstruction results}

\begin{figure}[htbp]
        \centering
            \includegraphics[width=0.48\linewidth]{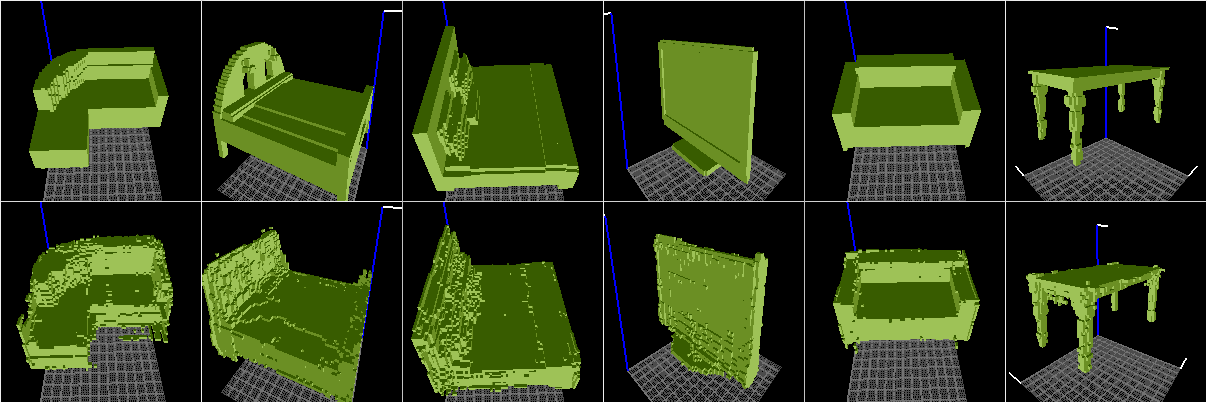}%
            \hspace{-0.25em}
            \includegraphics[width=0.48\linewidth]{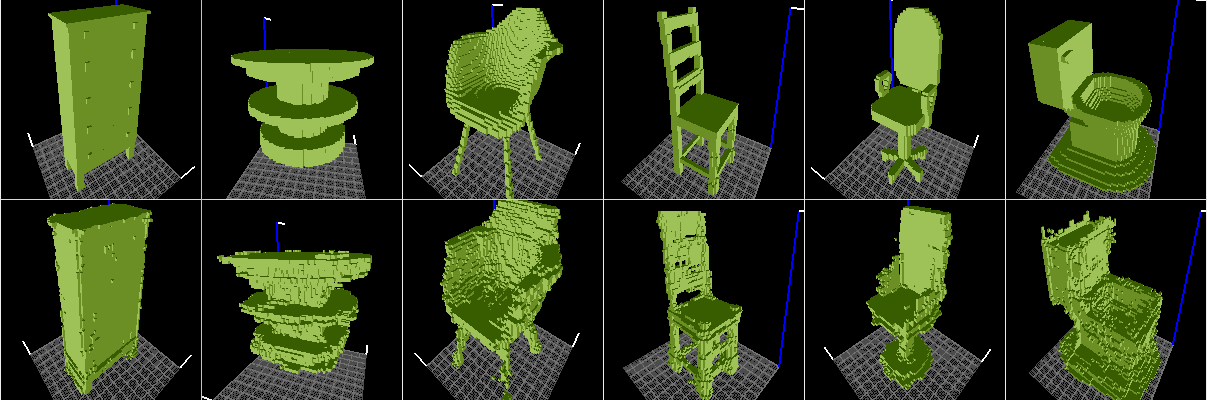}
        \caption
        {
            Examples of original voxelised objects (top) and their reconstructions (bottom) generated by the \gls{VAE} neural network.
            Objects taken from the
            ModelNet dataset \protect{\citep{3d_shapenets_modelnet_2015_7298801}}.
            %
        }
        \label{fig:origs_and_recons}
    \end{figure}

    Fig.~\ref{fig:origs_and_recons} illustrates
        3D object samples and their corresponding reconstructions
        generated by the network.
    The satisfactory quality of reconstructions suggests that
        the encoder network can generate descriptions of objects in a feature (latent vector) space, and that
        the decoder network can successfully reconstruct objects from descriptions generated by the encoder network.

    \FloatBarrier

    \subsection{Functional essence extraction results}

\begin{figure}[h]
        \centering
        \begin{subfigure}[t]{0.95\linewidth}
            \centering
            \includegraphics[width=0.8\linewidth]{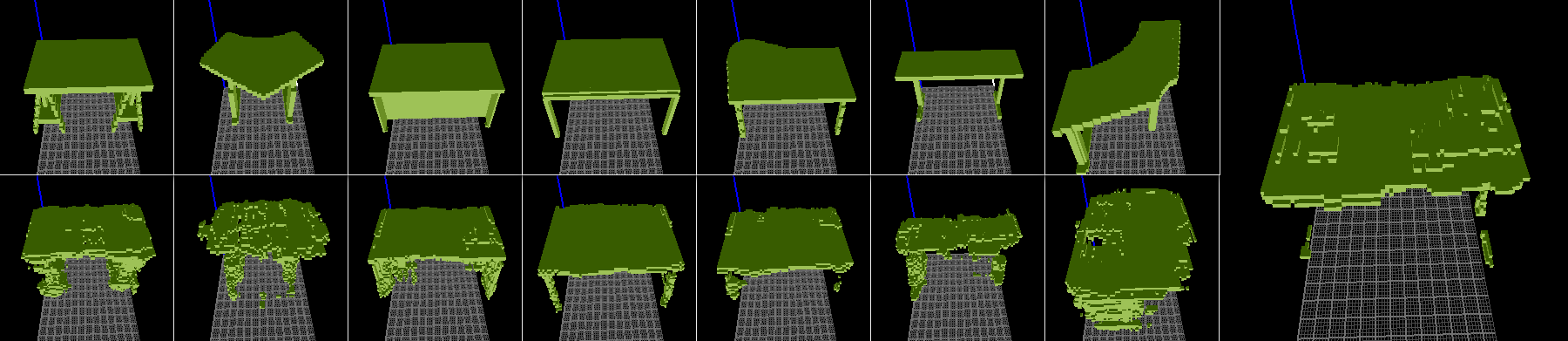}
          \caption{Sample tables (top), their reconstructions (bottom), and their common form features or \emph{functional essence} (right). The flatness feature was successfully extracted, which can be interpreted as providing the \emph{support-ability} of tables. Since supports differed in the samples, their were not included in the set of common shape features.}
          \label{subfig:essence_tables}
        \end{subfigure}
        \\ 
        \begin{subfigure}[t]{0.95\linewidth}
            \centering
            \includegraphics[width=0.8\linewidth]{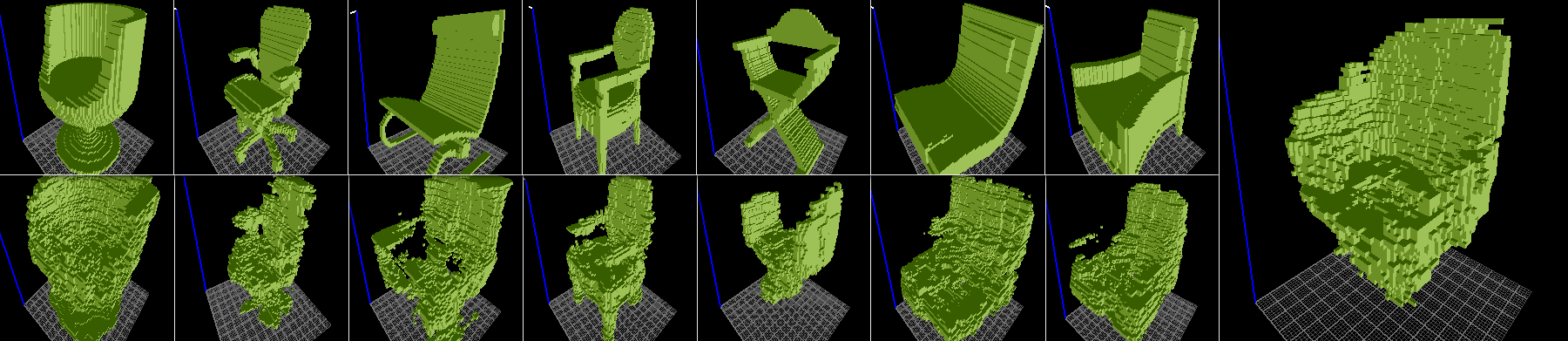}
            \caption{Sample chairs (top), their reconstructions (bottom), and their common form features (right). The seat and backrest are present in the set of common shape features, providing the \emph{sit-ability} and \emph{lean-ability} affordances. Since multiple chairs had armrests in the samples, these were also included in the set of common shape features.}
            \label{subfig:essence_chairs}
        \end{subfigure}
        \\ %
        \begin{subfigure}[t]{0.95\linewidth}
            \centering
            \includegraphics[width=0.8\linewidth]{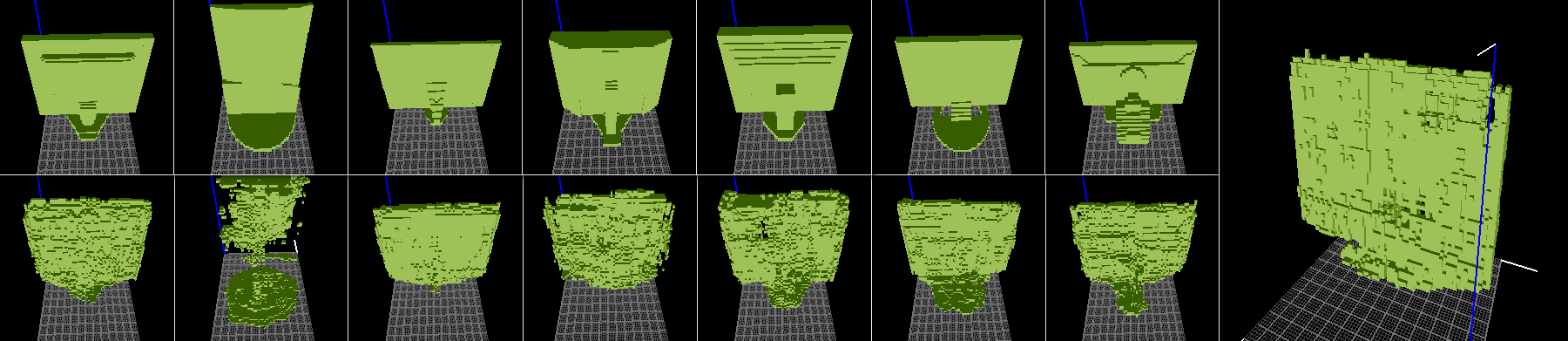}
          \caption{Sample monitors (top), their reconstructions (bottom), and their common form features (right). The flatness of screens was successfully identified as a common shape feature.}
          \label{subfig:essence_monitors}
        \end{subfigure}
        \caption
        {
            \textbf{Functional essences} extracted for
            (\subref{subfig:essence_tables}) tables,
            (\subref{subfig:essence_chairs}) chairs
            and
            (\subref{subfig:essence_monitors}) monitors.
            Objects taken from the
            ModelNet dataset \protect{\citep{3d_shapenets_modelnet_2015_7298801}}.
            Visualiser: viewvox \protect{\citep{binvox}}. 
        }
        \label{fig:functional_essences}
    \end{figure}

    Through the extraction of function essences of different object classes,
        we expected to identify forms that provide functionalities
        offered by those classes of objects.
    Fig.~\ref{fig:functional_essences} shows results on \emph{functional essence} extraction for
        tables, chairs, and monitors.
    Relevant features have been extracted, such as the flatness of tables providing \emph{support-ability}, the seats and backrests of chairs providing \emph{sit-ability} and \emph{lean-ability}, respectively.
    In the case of the chair object class,
        a considerable proportion of objects had armrests,
        which led to this feature becoming part of the functional essence.

    \FloatBarrier

    \FloatBarrier

    \subsection{Object combination results}

    The ability
        to extract a shape representation that constitutes the functional essence of a class,
        coupled with the ability to combine it with another object representation,
        makes it possible to extract and combine
        shape features that provide desired functionalities.
    It is worth noting that
        the proposed combination operator is non-commutative,
        meaning that the combination of two objects can
        generate different results,
        depending on the order of objects in the combining operation
        (i.e. which object is used as \emph{base object},
        and the order in which other objects are combined with it).

    \subsubsection{Sit-ability and wash-ability}

\begin{figure}[htbp]
            \centering
            \begin{subfigure}[t]{0.18\linewidth}
                \centering
                \includegraphics[height=0.7\linewidth]{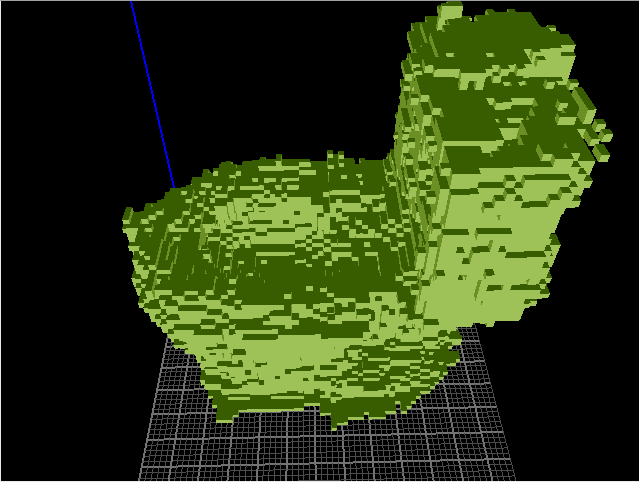}
            \caption{Top $50\%$ bathtub essence combined with a
            toilet essence base.}
            \end{subfigure}
            ~
            \begin{subfigure}[t]{0.18\linewidth}
                \centering
                \includegraphics[height=0.7\linewidth]{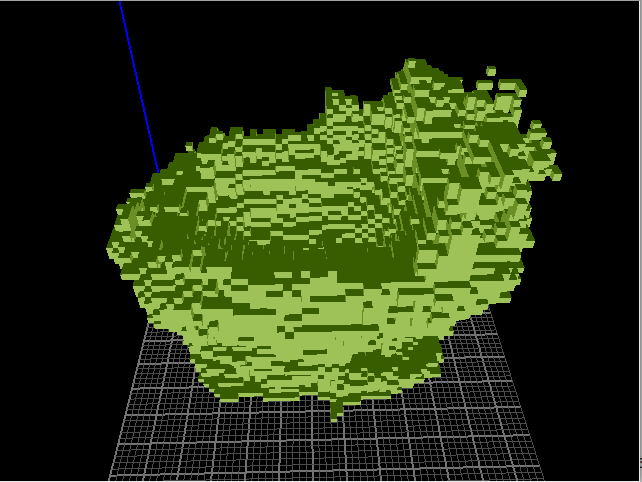}
                \caption{Top $60\%$ bathtub essence combined with a toilet essence base.}
            \end{subfigure}
            ~
            \begin{subfigure}[t]{0.18\linewidth}
                \centering
                \includegraphics[height=0.7\linewidth]{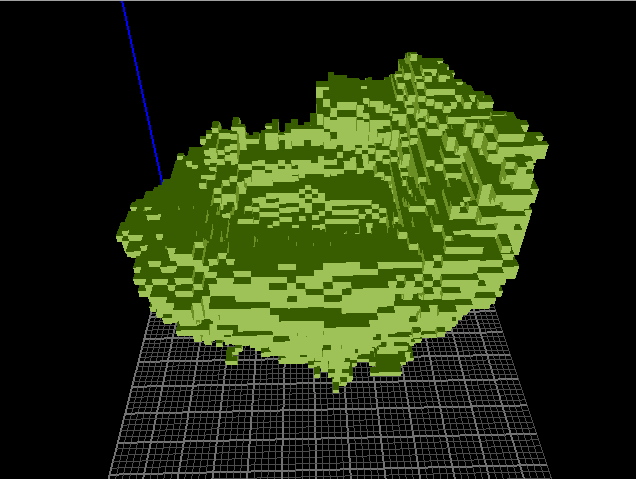}
            \caption{Top $70\%$ bathtub essence combined with a toilet essence base.}
            \end{subfigure}
            ~
            \begin{subfigure}[t]{0.18\linewidth}
                \centering
                \includegraphics[height=0.7\linewidth]{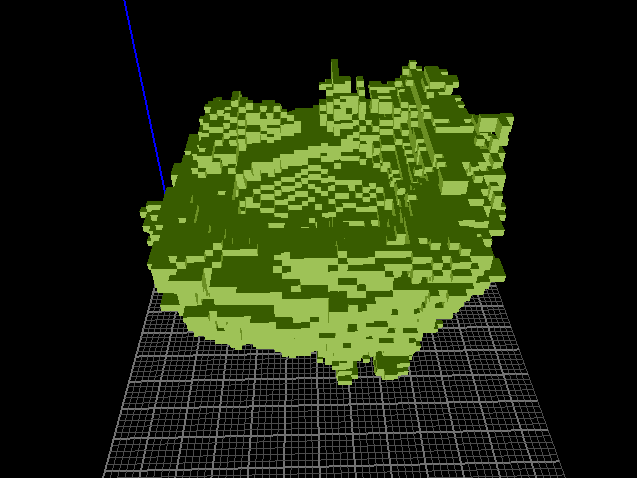}
            \caption{Top $80\%$ bathtub essence combined with a toilet essence base.}
            \end{subfigure}
            ~
            \begin{subfigure}[t]{0.18\linewidth}
                \centering
                \includegraphics[height=0.7\linewidth]{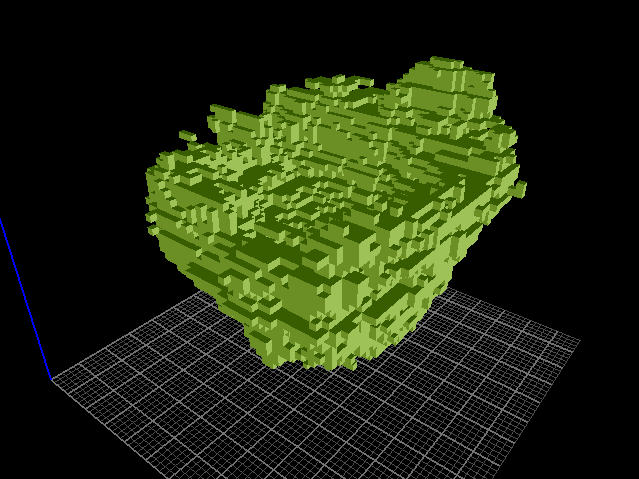}
            \caption{Top $95\%$ bathtub essence combined with a
            toilet essence base.}
            \end{subfigure}
        \\
            \begin{subfigure}[t]{0.18\linewidth}
                \centering
                \includegraphics[trim=20px 0px 15px 10px,clip,height=0.7\linewidth]{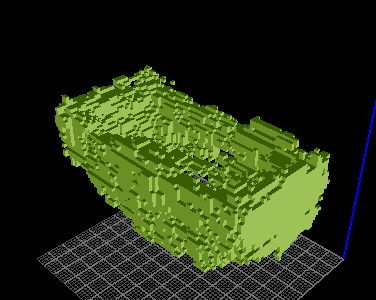}
                \caption{Bathtub essence base combined with top $50\%$ toilet essence.}
            \end{subfigure}
            ~
            \begin{subfigure}[t]{0.18\linewidth}
                \centering
                \includegraphics[height=0.7\linewidth]{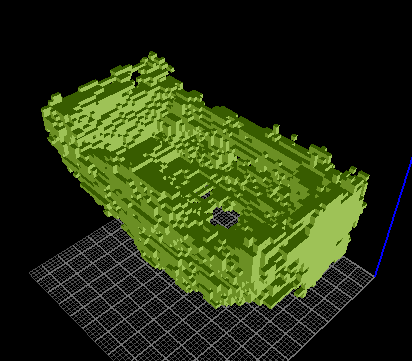}
                \caption{Bathtub essence base combined with top $60\%$ toilet essence.}
            \end{subfigure}
            ~
            \begin{subfigure}[t]{0.18\linewidth}
                \centering
                \includegraphics[height=0.7\linewidth]{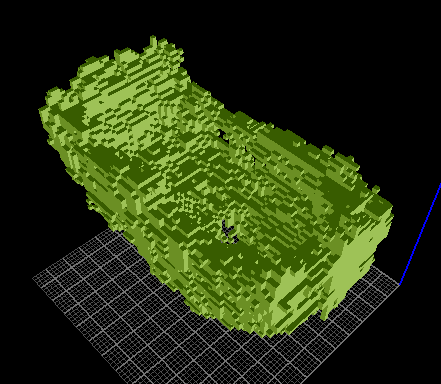}
                \caption{Bathtub essence base combined with top $70\%$ toilet essence.}
            \end{subfigure}
            ~
            \begin{subfigure}[t]{0.18\linewidth}
                \centering
                \includegraphics[height=0.7\linewidth]{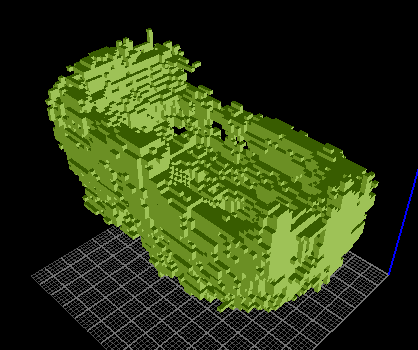}
              \caption{Bathtub essence base combined with top $80\%$ toilet essence.}
            \end{subfigure}
            ~
            \begin{subfigure}[t]{0.18\linewidth}
                \centering
                \includegraphics[height=0.7\linewidth]{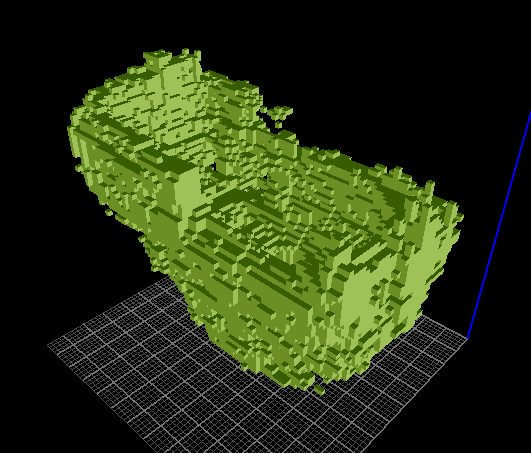}
              \caption{Bathtub essence base combined with top $95\%$ toilet essence.}
            \end{subfigure}
            \caption
            {
                \textbf{Object combination} results for bathtubs and toilets essences, using a \textbf{toilet essence base combined with a bathtub essence} (top),
                and a \textbf{bathtub essence base combined with a toilet essence} (bottom),
                both of which can be interpreted as a bidet.
                A gradual transformation is displayed (base functional essence combined with top-50\% to top-95\% of the second functional essence). From left to right, the combination looks less like a toilet (top) / bathtub (bottom) and more like a bidet.
            }
            \label{fig:bidet_bathtub_base_toilet_top}
        \end{figure}

    In this experiment,
        we have attempted to extract the \emph{sit-ability}
        and \emph{wash-ability}
        of toilet seats and bathtubs, respectively,
        in order to combine them into a new object
        providing both of these functionalities.
    The obtained results may be interpreted as bidet objects.
    The impact of the order in which objects are combined
        is visible in
        the two results
        displayed in Fig.~\ref{fig:bidet_bathtub_base_toilet_top}.
    The degree to which two object categories are combined
        can be controlled by varying the amount of information kept from each object description (i.e. the percentage of variables considered \emph{important} for an object description).

    \FloatBarrier

    \subsubsection{Wash-ability and support-ability}

\begin{figure}[htbp]
                \centering
                \begin{subfigure}[t]{0.23\linewidth}
                    \centering
                    \includegraphics[height=2.5cm]{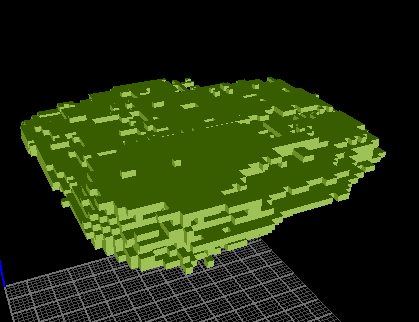}
                    \caption{Table functional essence, with its characteristic flatness providing \emph{support-ability}.}
                \end{subfigure}
                ~
                \begin{subfigure}[t]{0.23\linewidth}
                    \centering
                    \includegraphics[height=2.5cm]{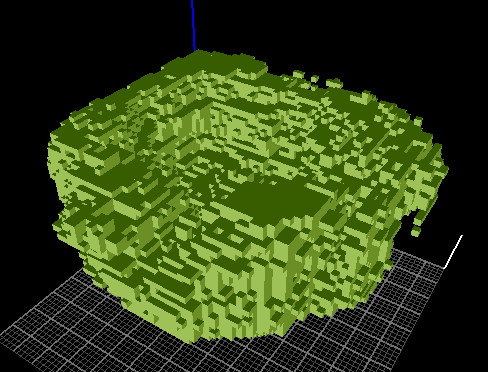}
                  \caption{Bathtub functional essence, providing \emph{wash-ability} with its convex shape.}
                \end{subfigure}
                ~
                \begin{subfigure}[t]{0.24\linewidth}
                    \centering
                    \includegraphics[height=2.5cm]{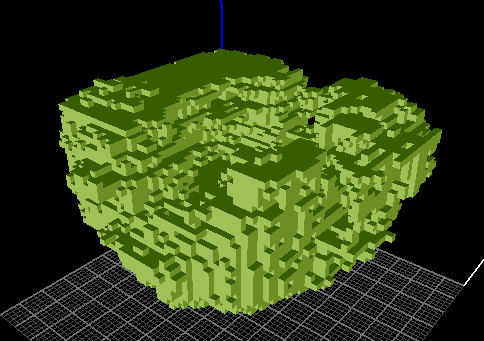}
                  \caption{A bathtub shape with a flat surface on top, 
                  providing both \emph{wash-ability} and \emph{support-ability}.}
                \end{subfigure}
                ~
               \begin{subfigure}[t]{.17\linewidth}
                    \centering
                    \includegraphics[height=2.5cm]{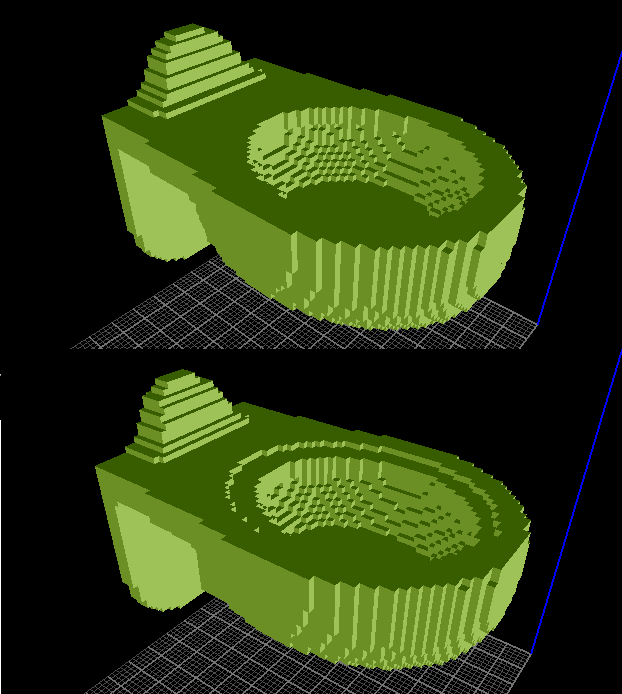}
                    \caption
                    {
                        Closest objects from the training set.
                    }
                    \label{subfig:similar}
                \end{subfigure}
                \caption
                {
                    Combining features of objects providing respectively \emph{wash-ability} and \emph{support-ability} into a novel object form, providing both functionalities.
                    (\ref{subfig:similar}) shows the two objects from the training dataset that are closest to our generated object, in terms of similarity of the activation values of the one-before-last layer of the decoder.
                }
                \label{fig:wash_support}
            \end{figure}

    This experiment displays the combination of \emph{support-ability} and \emph{contain-ability}  functionalities with the intent of creating something similar to a workdesk in a bathtub.
    The result is shown in Fig.~\ref{fig:wash_support}.

    \FloatBarrier

    \subsection{Quantitative results}

    We analysed the generated objects using 3 methods: (1) verification of affordance presence using state-of-art affordance detectors, (2) comparison of generated objects to most similar objects in the dataset, and (3) testing affordance presence in a physics simulation.
    These are detailed below. 

    \subsubsection{Affordance detectors}

    We attempted to identify the presence of the desired affordances (contain-ability,  support-ability)
    using affordance detectors developed by other groups.

    Sadly, we were not able to replicate the affordance detection results of \cite{myers2015affordance} on synthetic object images seen by a Kinect RGBD camera inside the Gazebo simulator.
    It failed to recognise the containability affordance in both standard objects like a bowl and a saucepan, and in generated ones.

\begin{figure}[htb]
          \centering
          \begin{subfigure}[t]{.48\linewidth}
              \centering
              \includegraphics[width=0.9\linewidth]{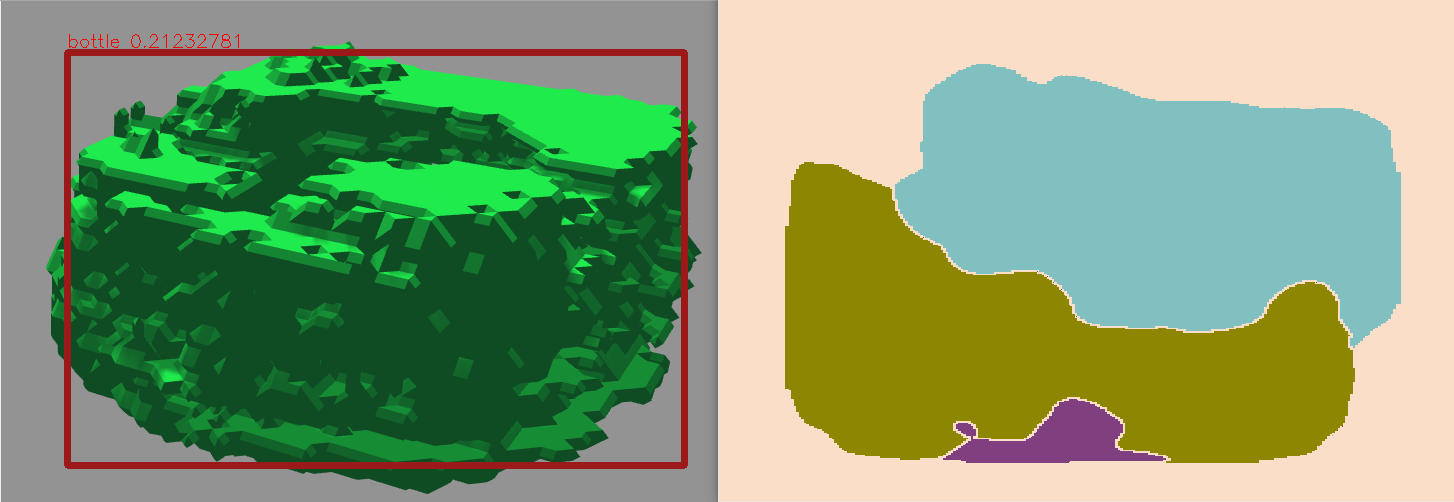}
            \caption
            {
            The AffordanceNet detector correctly identified
            \emph{support-ability} (in light blue) and \emph{wrap-grasp-ability} (in mustard colour),
            and incorrectly identified
            \emph{hit-ability} (in purple).
            }
            \label{subfig:affNet_a}
          \end{subfigure}
           ~
           \begin{subfigure}[t]{.48\linewidth}
               \centering
               \includegraphics[width=0.9\linewidth]{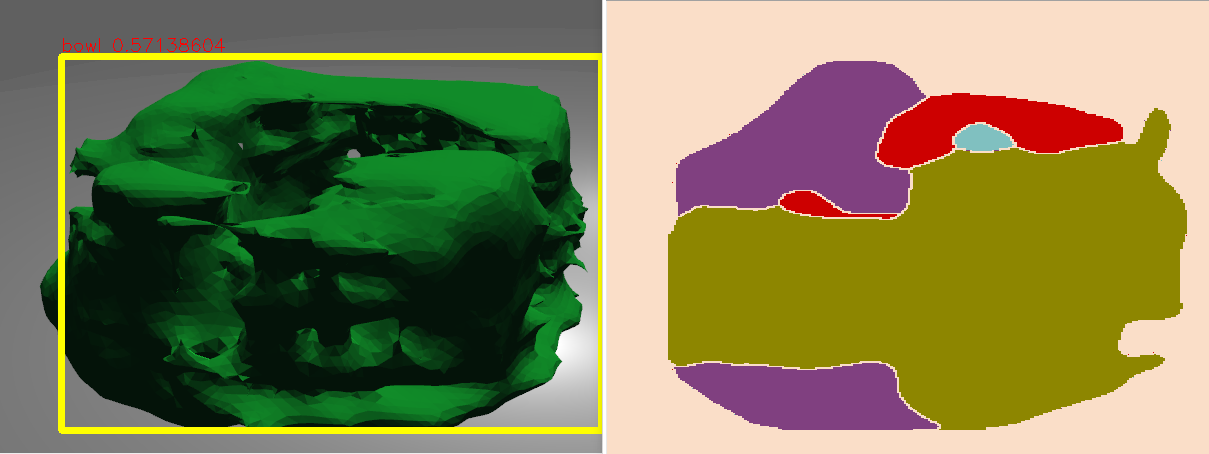}
             \caption{
             On a smoothed version of the object, and in different lighting conditions,
             it 
             correctly identified \emph{wrap-grasp-ability} (mustard),
             \emph{contain-ability} (red), although with imperfect segmentation.
             It incorrectly identified \emph{hit-ability} (purple) and \emph{support-ability} (light blue).
             }
             \label{subfig:affNet_b}
           \end{subfigure}
          \caption
          {
            Affordance detection results using the AffordanceNet \citep{do2018affordancenet}.
          }
          \label{fig:affNet_results}
      \end{figure}

    We also tried the affordance detector of \cite{do2018affordancenet}, called AffordanceNet.
    While it worked on objects viewed in simulation (including those of objects from the ModelNet40 dataset on which our network was trained), it had difficulties with recognising properly the affordances of generated objects (see Fig.~\ref{subfig:affNet_a}).
    We found experimentally that the failure cases for affordance detection were caused by the rugged surface of the object, and the fact that AffordanceNet was not trained on images of rugged objects.
    After applying Poisson smoothing to this object's surface, the detector correctly identified the presence of \emph{contain-ability}, although it still struggled to locate it properly (see Fig.~\ref{subfig:affNet_b}).

    \subsubsection{Most similar shapes in the training dataset to the ones generated}

    To ensure that the employed algorithm does not simply generate models by copying samples from the dataset, we compare the generated objects with the most similar samples from the dataset, based on the similarity of outputs of the one-before-last layer of the decoder. The result from Fig.~\ref{subfig:similar} confirmed that generated objects are distinct from the samples in the training set.

    \subsubsection{Affordance testing in simulation}

    To verify that the generated objects indeed provide the requested affordances, we developed some tests to execute in simulation.
    For this purpose, the generated voxelgrid model is transformed into a mesh using the \emph{marching cubes} method \citep{lorensen1987marching_cubes}, after which we compute its inertia matrix and create the \gls{SDF} file that allows to import it into the Gazebo simulator, using the Bullet physics engine. 

\begin{figure}[htb]
        \centering
        \begin{subfigure}[t]{.18\linewidth}
            \centering
            \includegraphics[height=2cm]{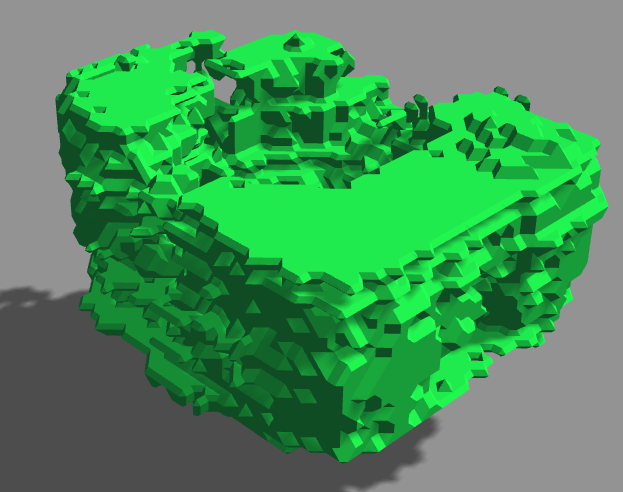}
          \caption
          {
            Perspective view of the generated object.
          }
          \label{subfig:bathub_perspective}
        \end{subfigure}
        ~
        \begin{subfigure}[t]{.18\linewidth}
            \centering
            \includegraphics[height=2cm]{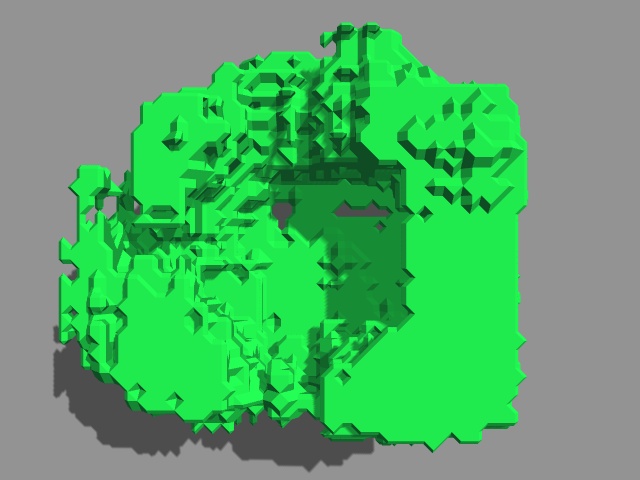}
          \caption{Top-down view of the generated object.}
          \label{subfig:bathtub_support_top_down}
        \end{subfigure}
        ~
        \begin{subfigure}[t]{.18\linewidth}
            \centering
            \includegraphics[height=2cm]{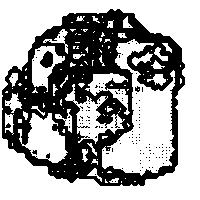}
          \caption{White pixels show locations with supportability.}
          \label{subfig:supportability_test}
        \end{subfigure}
        ~
        \begin{subfigure}[t]{.18\linewidth}
            \centering
            \includegraphics[height=2cm]{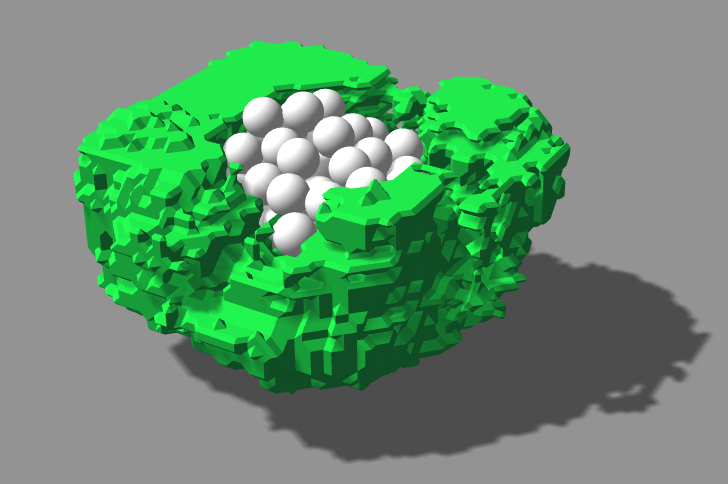}
          \caption{Containability test results with spheres.}
          \label{subfig:containability_test_spheres}
        \end{subfigure}
        ~
        \begin{subfigure}[t]{.18\linewidth}
            \centering
            \includegraphics[height=2cm]{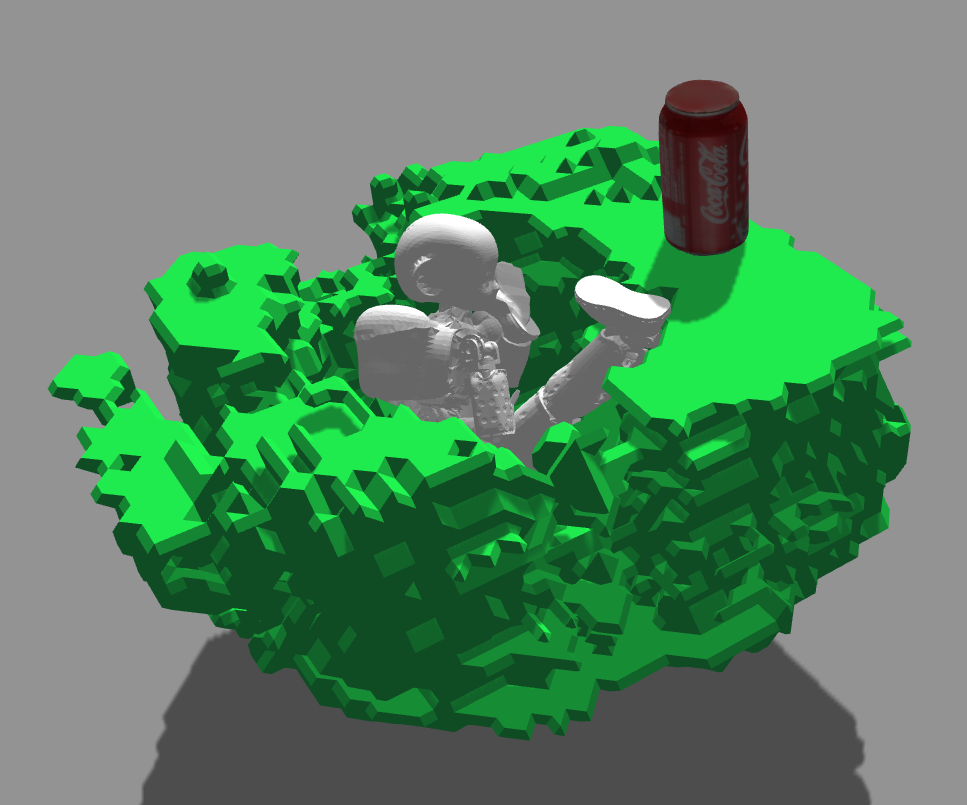}
          \caption{Containability of a humanoid robot in the bathtub.}
          \label{subfig:containability_test_icub}
        \end{subfigure}
        \caption
        {
            The generated bathtub-workdesk object in
            (\ref{subfig:bathub_perspective}) perspective view and
            (\ref{subfig:bathtub_support_top_down}) top-down view.
            (\ref{subfig:supportability_test}) shows the result of the \emph{support-ability} test, while
            (\ref{subfig:containability_test_spheres}) shows the result of the \emph{contain-ability} test.
            (\ref{subfig:containability_test_icub}) demonstrates that an iCub humanoid robot can fit inside the bathtub, and the Coca-Cola can illustrates the supportability of the workdesk.
        }
        \label{fig:supportability_test}
    \end{figure}

    To verify for supportability,
    we suspended the object into the air,
    and verified which of its regions 
    can support a stable object with a flat base, 
    by dropping from above from different (x,y) locations
    a 0.1 $m^3$ cube with mass 1 kg,
    and checking whether this had any impact on the (x,y) coordinates of its centroid.
    If only its z coordinate (altitude from ground) had changed,
    while the (x,y) coordinates remained the same, 
    then that location was marked as providing stability.
    On the contrary, if the region was not flat, the cube would tumble over, landing on (x,y,z) coordinates distinct from its initial ones.
    %
    Fig.~\ref{subfig:supportability_test} shows the obtained result.
    To verify containability, we dropped spheres into the object until they overflowed, and measured the ratio of the total volume of all spheres contained inside the object versus the volume of its bounding box (see Fig.~\ref{subfig:containability_test_spheres}).

    \FloatBarrier

    \subsection{Limitations}

        The proposed method currently has a set of limitations:
        %
        \begin{enumerate*}[label=(\roman*)]
        \item The method used for extracting functional essences from object categories, which employs \emph{averaging} out the gaussians describing the voxel locations, requires all samples in the dataset to be aligned.
        \item The combination method does not state
            if a solution to the posed problem does not exist
            (i.e.~if combining two different sets of affordances is possible).
        \item The different scales of objects are not taken into consideration when combining objects. Training the neural network on object models which are correctly sized relative to each other would solve this issue. However, it would require increasing the size of the input voxel cube to fit inside detailed descriptions of both small-scale objects (e.g.~spoons, forks, chairs) and large scale objects (e.g. dressers, sofas), which would also increase the training time.
        \item The network was trained on a dataset of mesh models
        (converted to voxelgrids), which are shells of objects, with a void interior.
        Thus, we implicitly trained the network to generate only hollow objects.
        %
        %
        \item Since the features describe the voxels mostly in the center of the bounding cube, combining two different feature descriptions makes them compete for the same center voxels in this bounding volume.
        Introducing an operator for spatially offseting some shape features would allow to construct composite objects.
        For instance,
            if we want to extract the \emph{sit-ability} and \emph{support-ability} from chairs and tables, respectively,
            in order to create something similar to a conference chair,
            it would be required to offset the table features 
            with respect to the chair features.
        \end{enumerate*}

\section{Conclusion and future work}
\label{sec:conclusion_future_work}

    We have presented a method for generating objects with desired functionalities, by first extracting a form-to-function mapping from a dataset of objects, and then manipulating and combining these forms through functionality arithmetic.
    The method relies on a neural network
        to extract feature-based 
        descriptions of objects.
    These descriptions
        allow shape manipulation and arithmetics
        in a latent feature space,
        before being transformed back into 3D object models.
    We then test the presence of desired affordances in a physical simulator, and with an affordance detector.


    In contrast to an ontology based approach,
        where modifications can be done deterministically,
        all the object shape manipulations are probabilistic in our case.
    Thus, generated inexact models may prove sufficient if regarded only as \emph{design suggestions}.
    However, a production-grade technology would require less noisy object-modeling results.
    We plan to employ a \gls{GAN} approach \citep{wu_GAN_NIPS2016_6096}, encouraging the network to generate objects with smooth surfaces similar to those of existing man-made objects.
    We also plan to implement a training procedure to
    encourage neurons in the latent layer to represent specific transformations (rotation, scale) following the approach of \cite{kulkarni2015deep}.

    Our models still lack
        information about
        materials from which objects are composed,
        their colors or textures (where necessary),
        and the articulations between subparts.
    Adding them 
        would make the approach much more practical. 

    In addition, instead of
        using a dataset containing an implicit mapping of form-to-function
        (as objects are categorised in classes),
        we intend to learn object functionalities/affordances
        automatically, by letting a robot interact autonomously
        with a set of objects.
    This is related to the currently active field of \emph{affordance learning} in robotics.
    Moreover, the use of 3D shape descriptors developed in this research
        will facilitate affordance learning and knowledge transfer
        in the case of autonomous robots.

    The source code will be made available 
    upon publication.
\subsubsection*{Acknowledgement}

This work was made possible with the help of a research grant offered by the \gls{IST-ID}.
Atabak Dehban is supported by the doctoral grant PD/BD/105776/2014 offered by the \gls{FCT}.
%
%
This research was also supported by a grant offered by the NVIDIA Corporation in the form of an NVIDIA Titan Xp GPU.
We thank Hugo Simão for his help in generating the images in Fig.~\ref{fig:pontoon_bridge_example}.
We also thank the anonymous reviewers who helped improve this manuscript.

\bibliographystyle{iclr2019_conference}
\bibliography{Bibliography/Bibliography.bib}


\end{document}